\documentclass[fleqn,10pt]{wlscirep}
\usepackage[utf8]{inputenc}
\usepackage[T1]{fontenc}
\usepackage{float}

\usepackage{makecell}
\usepackage{amssymb}
\usepackage{amsmath}
\usepackage{graphicx}
\usepackage[colorlinks=true, allcolors=blue]{hyperref}
\usepackage{tabularray}

\usepackage{changes}
\usepackage{xcolor}
\usepackage{orcidlink}

\title{Time Series Similarity Score Functions to Monitor and Interact with the Training and Denoising Process of a Time Series Diffusion Model applied to a Human Activity Recognition Dataset based on IMUs
}

\author[1,*]{Heiko Oppel \orcidlink{0000-0002-4410-2442}}
\author[1]{Andreas Spilz \orcidlink{0000-0002-9419-0663}}
\author[1]{Michael Munz \orcidlink{0000-0003-3427-3827}}
\affil[1]{AI for Sensor Data Analytics Research Group, Ulm University of Applied Sciences, Ulm, 89081, Germany}
\affil[*]{heiko.oppel@thu.de}


\begin{abstract}
Denoising diffusion probabilistic models are able to generate synthetic sensor signals.
The training process of such a model is controlled by a loss function which measures the difference between the noise that was added in the forward process and the noise that was predicted by the diffusion model.
This enables the generation of realistic data. 
However, the randomness within the process and the loss function itself makes it difficult to estimate the quality of the data. 
Therefore, we examine multiple similarity metrics and adapt an existing metric to overcome this issue by monitoring the training and synthetisation process using those metrics.
The adapted metric can even be fine-tuned on the input data to comply with the requirements of an underlying classification task.
We were able to significantly reduce the amount of training epochs without a performance reduction in the classification task.
An optimized training process not only saves resources, but also reduces the time for training generative models. 
\end{abstract}

\begin{document}
\flushbottom
\maketitle
\section{Introduction}
In machine learning classifier are used to identify pattern in samples to differentiate between multiple categories.
Often the data basis is either missing samples from specific categories as it can be a time or cost consuming process or the data is of poor quality.
In such cases Denoising Diffusion Probabilistic Models (DDPMs) have emerged as powerful generative tools to increase the sample space with meaningful representatives, for example in domains such as computer vision \cite{imageclassificationWithSyntheticData} or time series \cite{timegrad}.
Those samples are then used to achieve better results in the classification task. 
Therefore, the synthesized data has to increase the variation of the dataset while also retaining the main information from the activity.
\newline
The training of such a DDPM is based on the maximization of the log-likelihood, so, that the generated sample distribution matches the one from the real data \cite{ddpm_ini}.
In order to achieve this, the loss function of the DDPM is defined as the mean squared error between the noise, that was estimated by the U-NET and the noise that was used in the forward process of the diffusion model.
This ultimately guarantees the generation of synthetic data in the forward process.
Though, it is not possible to estimate the quality of the generated data with this loss function or the resemblance to the real sequence.
In image generation, to assess the quality of the generated images one can rely on human raters as Best-Rowden and Jain \cite{image_similarity_human_rater} did.
With time series data, this is not feasible.
Among others, some studies rely on a qualitative analysis by using decomposition methods like t-SNE or analyzing the probability density functions \cite{tsne_and_pdf} between real and generated signals.
The disadvantage of those approaches is their requirement for visual confirmation.
It is not possible to reduce the similarity information to a single value. 
Another possibility is the usage of a discriminative score \cite{discrimnative_score}.
For this, a neural network is trained to differentiate between real and generated signals.
Though, this is a time consuming process and depending on the dataset the network architecture has to be adapted.
The Context Fréchet Inception Distance (FID) \cite{contextFID} is another approach that relies on the usage of a neural network model. 
In this case, the TS2Vec \cite{ts2vec} model is used.
It is able to map each time step of a time series to a contextual representation by learning a non-linear embedding function.
Some studies did also rely on similarity score functions to estimate the similarity between real and generated signals.
So did Liao et al. \cite{corrscore_ini} by calculating the absolute error of the auto-correlation estimator. Suh et al. \cite{suh2024timeautodiff} applied a similar methodology, but used the pair-wise column correlations among other evaluation methods.
Finally, some studies use an underlying classification or regression task to objectively estimate the quality of the generated signals \cite{NEURIPS2019_c9efe5f2}, \cite{suh2024timeautodiff}, \cite{imudiffusion}.
They evaluate the separability of the classifier with and without the addition of synthetic data. \newline
To sum it up, in the literature, there exist several approaches to evaluate the quality of the generated data, though, they were either not used to monitor the training progress of a diffusion model or are not suitable to do it.
Therefore, we not only introduce a similarity score, but also integrate it in the training and denoising process of a time series diffusion model by estimating the models ability to generate comparable signals. 
We do this by first calculating the power spectral density (PSD) of the signals and then estimate the similarity score using several similarity score functions separately.
The score functions we examine are the root mean squared error (RMSE), the Pearson correlation coefficient, the cosine similarity, and a novel adaptation of the global alignment kernel (GAK).
To the best of our knowledge, this is the first time that time series similarity score functions are used for monitoring the training and denoising progress of a diffusion model. 
So far, monitoring the training progress has been done by relying on the loss value alone. 
Using the score functions allow us to significantly reduce the amount of training epochs whilst also increasing the quality of the generated signals. \newline
\newline
This work has the following main contributions:
\begin{itemize}
    \item We propose a new similarity score to the domain of evaluating synthetically generated time series signals.
    \item We integrate similarity score functions in the training progress of a DDPM to reduce the amount of training epochs whilst maintaining or even improving the quality of the generated sequences.
    \item We use the similarity score functions to reduce the amount of denoising steps without decreasing the quality of the generated signals for the underlying classification task.
\end{itemize}

\section{Methods}
This is a follow-up study based on the work of Oppel et al. \cite{imudiffusion}.
The processing of the data, the choice of the classes, the DDPM model (IMUDiffusion) and classifier configuration are explained in detail in this study.
We will still provide a short explanation of those methods, but for more information, please refer to the original publication.

\subsection{Human Activity Recognition Dataset}
For this study, we used a human activity recognition dataset on the basis of Inertial Measurement Units (IMUs) introduced by Banos et al. \cite{displacement_dataset}. The original aim was to analyze the effect of IMU displacement.
They recorded $17$ participants performing $33$ activities from which we chose four: Walking, Running, Jump Up and Cycling.
As some of the participants did not participate in all activities, we reduce the pool of participants to those $12$ which performed all activities.
For a better readability when addressing single participants, we will further address them as PID $x$ (participant with the id $x$, $x \in {1,...,16}$).
Furthermore, we only used the IMU located on the right thigh with the ideal placement setup.
This guarantees comparable movement pattern along the IMU axes.

\subsection{Signal Processing}\label{section:signal_processing}
The activities were recorded with a sample rate of $50\,Hz$. 
We further sequenced the data with a sliding window width of $160$ time steps and an overlap of $40$ time stemps.
Those signals are transformed into the frequency domain using a short time fourier transform (STFT) using a window size of $22$ and an overlap of $20$. 
Windowing has been done using the Hanning function. The frequency domain signal is then used as input into the diffusion model.

\subsubsection{Power Spectral Density}
The main goal of synthesizing data is to add variation to the dataset while retaining the main information from the activity.
Comparing sequences in the time domain may either suggest to use sequences that are highly similar, hence, not increasing the variance within the dataset or, even worse, it can lead to the assumption, that sequences are fairly dissimilar while having the key information of the activity, yet, deviate from the real sequences.
To address this issue, we estimate the signals' power spectral density (PSD) using Welch's method \cite{welch_psd}.
This method estimates the PSD by first separating the signal into $K$ windowed subsequences
\begin{equation}
    x_{\omega, k} = \omega x_k \text{ , with k=0,1,...K-1}
\end{equation}
where $\omega$ represents the window function.
For each subsequence, the periodogram $P_{x_{\omega, k}, M}(\omega)$ is then calculated
\begin{equation}
    P_{x_{\omega, k}, M}(\omega) = \frac{1}{M} \lvert\sum_{m=0}^{M-1} x_{\omega, k}(m)\cdot e^{-j2\pi m/M} \rvert ^2
\end{equation}
where $M$ denotes the sequence length of each subsequence.
Finally, by taking the average over all periodograms we get the power spectral density
\begin{equation}
    \hat{S}_x(\omega) = \frac{1}{K}\sum_{k=0}^{K-1}{P_{x_{\omega, k}, M}(\omega)}
\end{equation}
Using this approach removes the temporal dependency in the course of the sequence.
The idea behind this is to focus on the main characteristics in the signal that represents the activity independent of the location in time.

\subsection{Similarity Metrics}
\subsubsection{Class-Optimized Global Alignment Kernel}\label{section:GAK}
\begin{figure}
    \centering
    \includegraphics[width=0.7\linewidth]{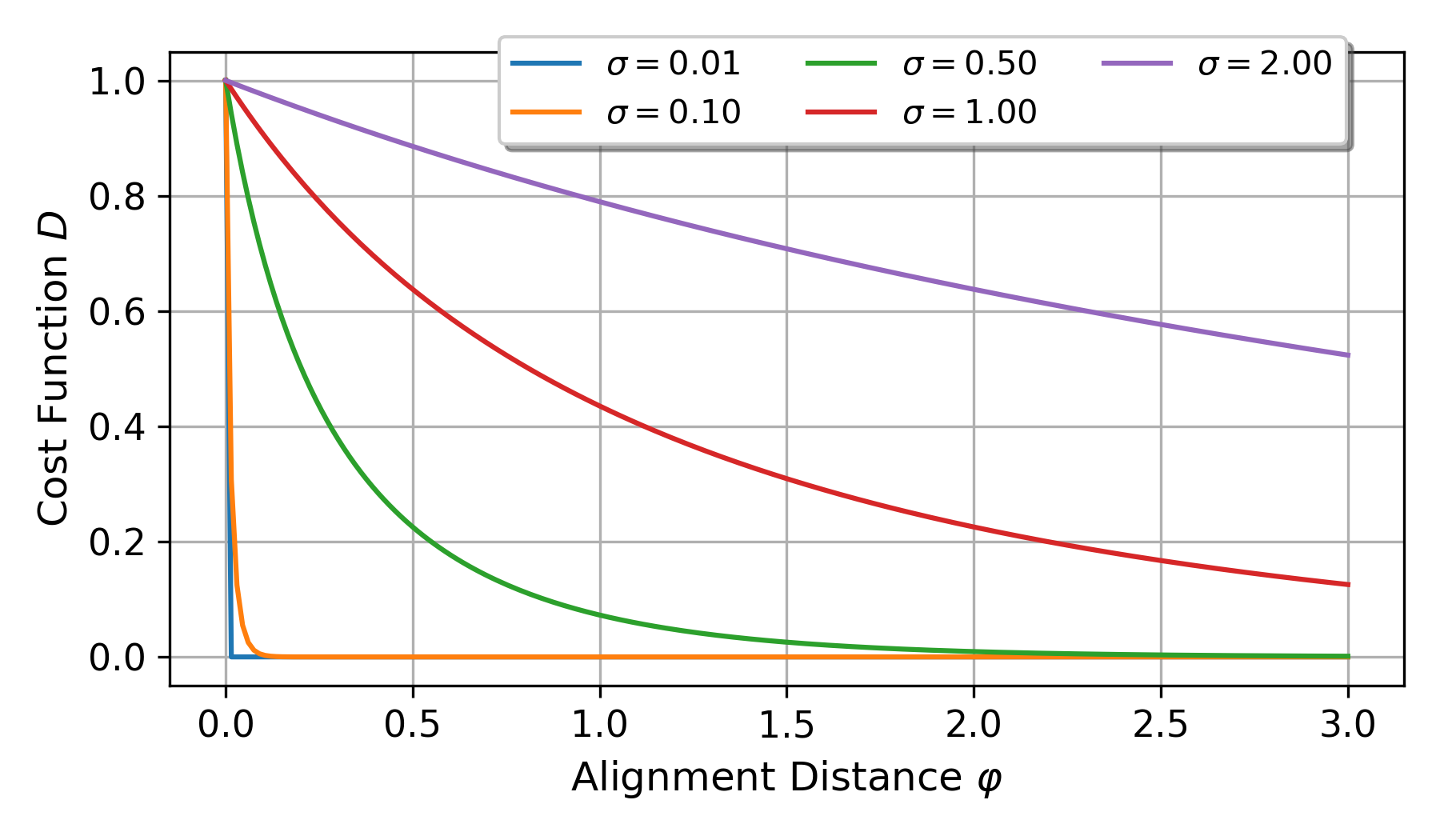}
    \caption{Influence of the scaling factor $\sigma$ on the cost function $D$.}
    \label{fig:sigma_einfluss}
\end{figure}

Global Alignment Kernel (GAK) $k(x, y)$ is an approach to map a sequence $x$ onto another sequence $y$.
As stated in~\cite{fast_global_alignment}, it exponentiates the soft-minimum of all alignment distances and is defined as 
\begin{equation}\label{eq:k}
    k(x,y)=\sum_{\pi \in \mathcal{A}(n, n)} e^{-D_{x,y}(\pi)} \text{,}
\end{equation}
whereas $\pi$ being an alignment path, $\mathcal{A}(n, n)$ the set of all alignments between the two sequences $x$ and $y$ of length $n$ and $D$ is the cost of the alignment $\pi$.
An alignment path is a sequence of index pairs which best map the sequences $x$ and $y$ onto each other.
The cost $D$ is defined by equation~\ref{eq:D} and its exponentiation bounds each element to $[0, 1]$.

\begin{equation}\label{eq:D}
    D_{x,y} = d(x,y)-\ln{(2-\exp{(d(x,y)})} \text{, with } d(x,y)=-\frac{\varphi(x,y)}{2\sigma^2} \text{ and } \varphi(x,y)=\sqrt{(x-y)^2}
\end{equation}
Each operation in calculating the cost function is an element-wise operation.
The scaling factor $\sigma$ is responsible for the scaling of the distance function, and, hence, on the cost function $D$, see Figure~\ref{fig:sigma_einfluss}.
In summary, by increasing $\sigma$, the cost function approaches its limit value $0$ slower.

Finally, we normalize the global alignment kernel $k(x,y)$ according to equation~\ref{eq:gak_normalization}.
\begin{equation}\label{eq:gak_normalization}
    \vartheta(x,y) = \frac{k(x,y)}{\sqrt{k(x,x) \cdot k(y,y)}} \in \mathbb{R}:\vartheta(x,y)\in [0, 1]
\end{equation}

\begin{figure}
    \centering
    \includegraphics[width=0.9\linewidth]{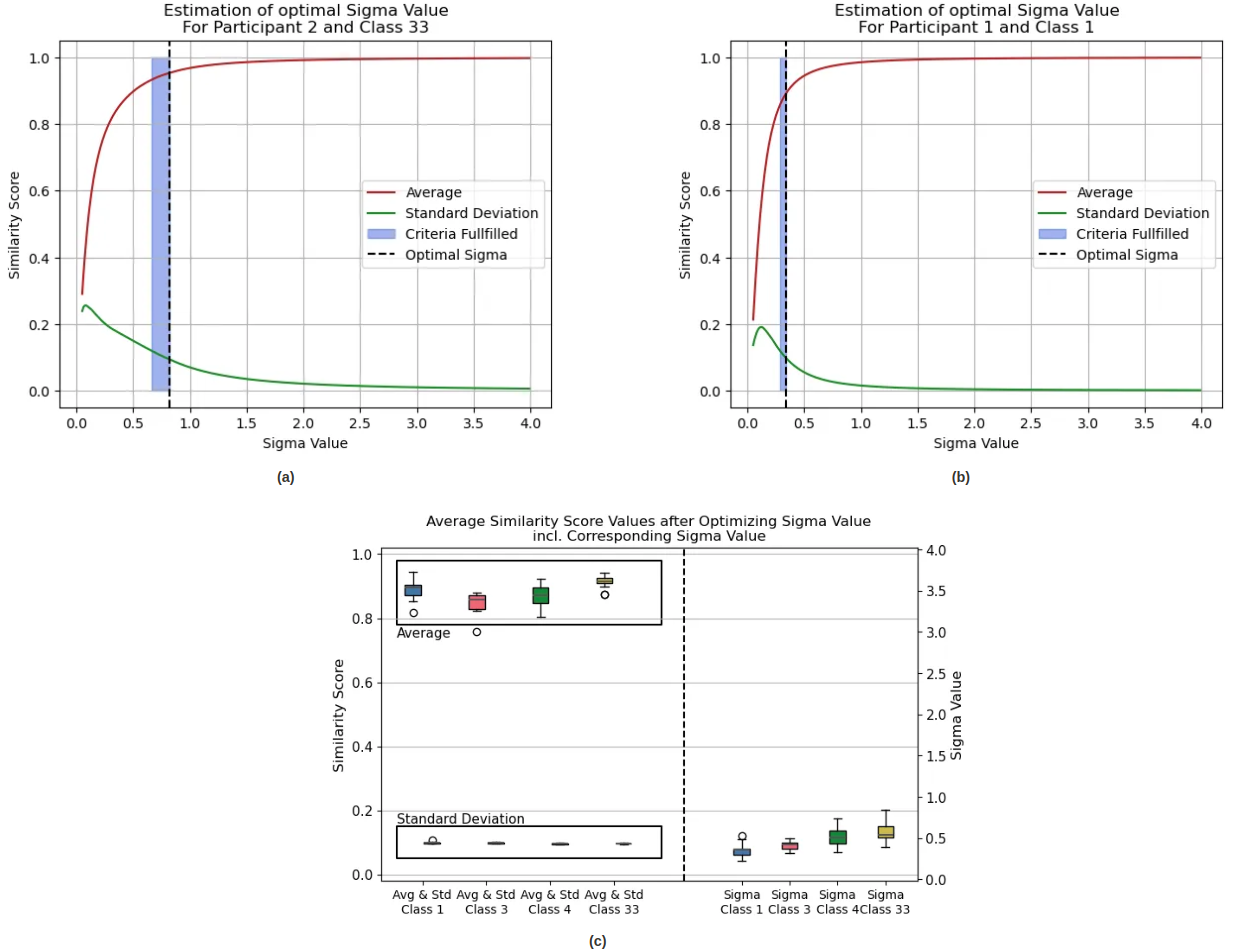}
    \caption{Visual identification of the optimal sigma value for a given train and validation set: (a) for the Cycling activity performed by the participant with the ID 2; (b) for the Walking activity performed by the participant with the ID 1; (c) a summary over all participants by activity including the average and standard deviation of the similarity scores across all participants.}
    \label{fig:optimal_sigma}
\end{figure}

\textbf{Estimation of the optimal sigma value}\newline
The GAK is directly dependent on the scaling factor $\sigma \in \mathbb{R}:\sigma>0$.
It is a sensitive parameter responsible for the degree of selectivity of the similarity between two sequences.
A high degree of selectivity means that small variations in the data are able to change the value of the GAK significantly.
The lower the value, the higher the degree of selectivity.
Cuturi et al.~\cite{fast_global_alignment} suggest to calculate the scaling factor based on the median distance between various timesteps across the two time series and scale it.
It is even possible to use a multiple of the scaled median distance.
We evaluated their approach by calculating the GAK between sequences from our training and validation set, which should have a high degree of similarity.
Though, it lead to an average sigma value of $7.15\cdot10^{-4} \pm 4.87\cdot10^{-4}$, equivalent to a high degree of selectivity, and hence, made it not usable for our concept.
Therefore, we change the approach of estimating the optimal scaling factor.
As previously mentioned, we assume a high similarity between sequences from the training and validation set.
So, we perform an optimization by calculating the maximum of the average GAK value across all sequence pairs under the condition, that the standard deviation is in the range $[0.09, 0.12]$.
Due to the cyclic behavior of the activities, we assume a high similarity between data in the training and validation sets.
Therefore we analyzed the similarity between those sets in combination with the similarity score and finally decided on the previous mentioned range.
As this is a subjective assessment, it requires knowledge about the underlying dataset.
The mathematical formulation is as follows:

\begin{equation}\label{eq:optimal_sigma}
    \begin{aligned}
        C_{GAK}=\max(\bar\vartheta(x,y)),\\
        \text{subject to } \hat{\sigma}_{\vartheta}\in[0.09, 0.12],
    \end{aligned}
\end{equation}
with $\bar\vartheta(x,y)$ and $\hat\sigma_{\vartheta}$ being the average and standard deviation of the GAK values.
The range for the standard deviation was defined after analysing the GAK value of the most similar sequences for various $\sigma$-values.
Unfortunately, this still remains a partially subjective analysis.
This adapted GAK metric will be further referenced as the class optimized global alignment kernel (C-Opt GAK).\newline
A visual representation of the identification of the optimal sigma value is presented in Figure~\ref{fig:optimal_sigma}.
It visualizes the similarity score over a range of sigma values.
The blue range defines the area in which the criteria according to equation~\ref{eq:optimal_sigma} is fulfilled. 
The red and green curves describe the average and standard deviation over the most similar sequences. 
Depending on the input sequences, the sigma value is able to change the interpretation of the GAK value, compare the Figures~\ref{fig:optimal_sigma} (a) and (b).\newline
A summary of all calculated sigma values from all participants is presented in Figure~\ref{fig:optimal_sigma} (c). 
They range from $0.1$ in case of the Walking class up to almost $1.0$ for participants performing the Cycling activity.
Each class requires a specific range of sigma values to meet the criteria defined in equation~\ref{eq:optimal_sigma}.
In addition to the sigma values, Figure~\ref{fig:optimal_sigma} (c) visualizes the average and standard deviations of the C-Opt GAK values between the training and validation set. 
It is further separated by the individual classes.
The standard deviations $\hat{\sigma}_{\vartheta}$ show little variations as they are strictly limited by the condition.
The dispersion on the mean values $\bar{\vartheta}$ on the other hand differ between the activities. \newline

\subsubsection{Comparison Similarity Metrics}
We choose three time series similarity metrics to evaluate our C-Opt GAK metric against - the Cosine similarity, the Pearson Correlation Coefficient and the Root Mean Squared Error (RMSE).
Each metric is used to calculate the similarity scores in the time domain in addition to analyzing the similarities of the signals power spectral densities.\\
The Cosine similarity $s_c$ between two sequences $x=(x_1, ..., x_n)$ and $y=(y_1, ..., y_n)$ is calculated by taking the dot product between two sequences and additionally norming it using their magnitudes $||x||$ and $||y||$:
\begin{equation}
    s_c(x,y) = \frac{xy}{||x||\cdot ||y||}.
\end{equation}
The second similarity metric is the Pearson Correlation Coefficient $s_p$:
\begin{equation}
    s_p(x, y) = \frac{n\sum_i x_iy_i - (\sum_i x_i \cdot \sum_i y_i)}
    {\sqrt{(n\sum_i x_{i}^{2} - (\sum_i x_i)^2) \cdot (n\sum_i y_{i}^{2} - (\sum_i y_i)^2)}} \text{ ,with i=1,...,n}
\end{equation}
It is calculated as follows:
\begin{equation}
    s_r(x,y) = \sqrt{\frac{\sum_i(x_i - y_i)^2}{n}}
\end{equation}
Both Cosine similarity and Pearson Correlation Coefficient are in the range of $[-1, 1]$, the score value calculated by the RMSE is in the range of $	\mathbb{R}$.

\subsubsection{Visual Analysis of Exemplary Sequences between the Similarity Metrics}
\begin{figure}
    \centering
    \includegraphics[width=0.8\linewidth]{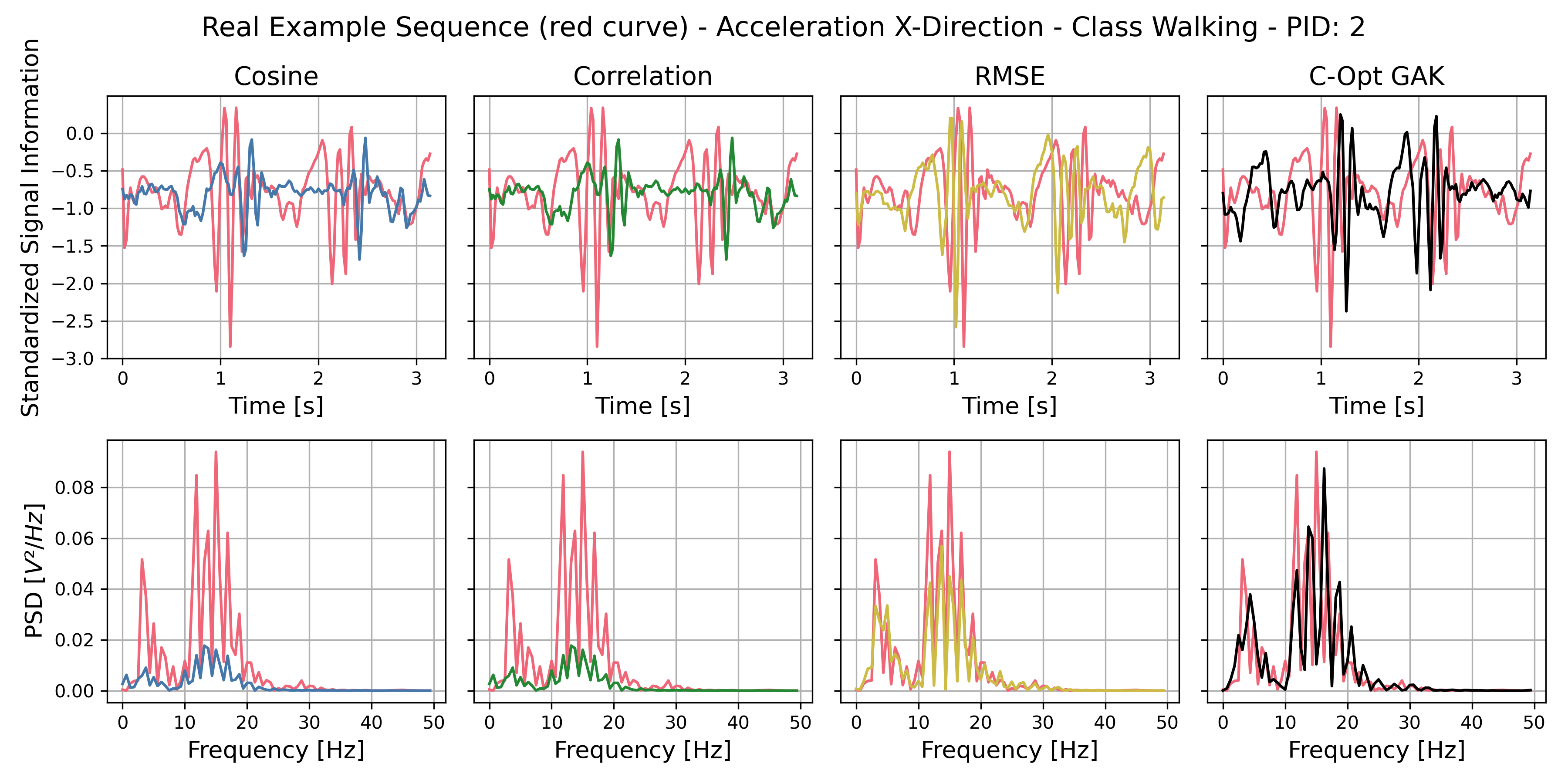}
    \caption{Visual comparison between a sequence from the training and validation set. 
    The sequence from the training set was randomly chosen and the sequence from validation set was chosen based on the similarity metric. 
    It is the sequence that resembles the training sequence the most according to the respective similarity metric. 
    The sequence from the training set is visualized in red.
    The respective sequence from the validation set is visualized in a different color depending on the similarity metric that led to choosing the respective sequence.
    The top row visualizes the sequences in the time domain and the bottom row their respective power spectral density.}
    \label{fig:most_similar_examples}
\end{figure}

Figure~\ref{fig:most_similar_examples} visualizes an example sequence showing the acceleration in x-direction from PID 2 of the Walking class once in the time domain and once its power spectral density (red curve in the figures).
We have visually compared this sequence with the most similar sequence from the validation set according to all four metrics.
To determine the most similar sequence, we calculated the similarity scores individually across all sensor axes and then averaged them.
The calculation was done between the power spectral densities of the sequences.
The Cosine and Correlation metric chose identical sequences, whereas the RMSE and C-Opt GAK approaches chose different sequences.
The choice for identical sequences between the Cosine and Correlation metric was observed across all classes. 
We have therefore decided not to consider the Correlation metric further in our analysis.
One of the disadvantages of using RMSE as a similarity score is the lack of interpretability of the score value itself.
The only assumption that can be made is the following: the lower the score value, the higher the similarity between two sequences.
Visually, both the sequences chosen by the RMSE as well as the C-Opt GAK show high similarities towards the sequence from the training set.
Therefore, we also excluded the RMSE score in our further analysis.

\subsection{Denoising Diffusion Probabilistic Model}
The IMUDiffusion model is a diffusion model specifically designed for synthesizing time series sequences based on multiaxial IMUs. 
It was first introduced by Oppel et al.\cite{imudiffusion} which showed the effectiveness of the generated sequences by improving the underlying classification task of separating human motion activities.
The model description can be found in their paper.
They trained the diffusion model for $4500$ epochs, which will be the reference for this study.\newline
For the noise scheduler, a linear scheduler was applied.
It was adapted to the multi-sensor problem by choosing separate diffusion rates per sensor.

\subsection{Classifier}
Like the diffusion model, we rely on the same classifier architecture introduced by Oppel et al. \cite{imudiffusion}.
This allows for a comparability between the results to analyze the effectiveness of the similarity score.
The classifier is composed of a convolutional neural network that convolves the input only along the time dimension.
We use two convolutional layers with a kernel of size $c_{kernel}=(1\times5)$ and $2$ filter each, followed by a Max-Pooling layer to reduce the dimensionality along the time by two. 
After the Max-Pooling layer an additional convolutional layer with $4$ filter and the same kernel size as the previous convolutional layer was used. 
Finally, the last part of the network consists of three fully connected layers with $128$, $32$ and $16$ neurons respectively. 
Each fully connected layer is followed by a ReLU activation layer and a dropout layer with $p=0.3$.
The last layer of the network consists of four neurons followed by a softmax layer.

\subsection{Experiments}
The similarity metrics Cosine Similarity and the C-Opt GAK are used to monitor the training and the denoising process of the DDPM.
Both metrics are used to compare the power spectral densities of the sequences.
Additionally, the Cosine similarity was also used to compare sequences in the time domain. \newline
In each experiment, we trained the DDPM and the classifier using the Leave-one--subject-out Cross Validation (LOSOCV) method.
Each participant was once excluded from the training and validation set and only used for testing.
As we have $12$ participants in total, we trained $12$ classifier models and evaluated the results separately.
The same methodology has been used to train the DDPM.
Though, the training of the DDPM was additionally separated by the classes to guarantee a unique label for the synthetic sequences.
This results into a total of $48$ DDPMs.

\subsubsection{Monitoring the DDPM Training}\label{section:monitoring_ddpm_training}
By monitoring the training progress of the DDPM using the similarity score functions, we are able to estimate the quality of the synthetic sequences at any desired epoch. 
To do this, we are denoising a batch of $128$ randomly normal distributed sequences in the frequency domain for the full $3000$ denoising steps after the specific epochs.
Though, as this is a time consuming process, we reduce the amount of monitored epochs to every $50^{th}$.
Now, the termination criteria between the C-Opt GAK and Cosine metrics vary.
Using the Cosine metric for monitoring the training process, we search for a local maxima of this metric between real training sequences and the batch of synthetic sequences.
Additionally, as the score value can be volatile, we keep training for another $100$ epochs including two monitoring steps, to ensure that an optimum has been reached.
The C-Opt GAK method allows us to be more specific with the criteria for terminating the training process.
By optimizing the scaling factor $\sigma$ using the real training and validation sets, we also estimate the range of the similarity score.
Therefore, we expect the similarity score between real training sequences and synthetic sequences to be in the same range. 
In practical terms, we require that at least $25\%$ of the similarity scores to be in the range.
If both criteria were met, we stop the training process.

\subsubsection{Monitoring the Denoising Process}\label{section:monitoring_denoising}
The scheduler is responsible for controlling the denoising process.
Initially, we set the number of denoising steps to $3000$.
With the help of the similarity scores, we are able to monitor this process and estimate the quality of the synthetic sequences by comparing them against the real training sequences.
We used the information of the similarity scores to stop the denoising process as soon as the optimal quality of the synthetic sequences was reached. 
Again, we allow two additional monitoring steps to guarantee that the local optimum was reached.
Therefore, if in two consecutive steps the similarity score drop, we stop the denoising process. 
Again, this is a very time consuming process if every denoising step is monitored. 
Therefore, we monitored only every $30^{th}$ step.

\subsubsection{Training Sets for the Classification Task}
In order to objectively evaluate the quality of the synthetic sequences, we add those sequences to the training set for classifying the four activities Walking, Running, Jump Up and Cycling.
Overall, we compare $9$ training sets against each other that have been used to train a neural network classifier with identical architecture and initial weights.
First of all, we have the two baseline sets - namely the "Full-Set" and the "2 Sample Set".
The Full-Set comprises $80\%$ of the available data from $11$ participants.
The remaining $20\%$ from those participants are used for validating the classifiers performance.
Finally, the left-out participant was used for testing.
Therefore, the test set was always identical, independent of the training sets.
In case of the 2 Sample Set, the training data comprises $2$ randomly chosen samples out of the real samples per participant, leading to a total of $22$ real samples in the set.
The same amount but different samples, were chosen for the validation set.
The final baseline set is the Full DDPM Set.
It consists of synthetic sequences which have been generated with the IMUDiffusion model without the usage of similarity metrics to monitor the training and synthetization process.
Meaning, the IMUDiffusion model has been trained for $4500$ epochs and the sequences have been denoised for $3000$ steps.
The same real sequences from the 2 Sample baseline set were used to train the diffusion models. \newline
The results obtained by the classifier with the baseline sets serve as a reference against the results obtained from training the classifier with a different training set that contains synthetic sequences which have been generated with the help of the similarity metrics.
Those two metrics were the C-Opt GAK and Cosine similarity and were either applied to monitor the training of the IMUDiffusion model or its denoising process.
Depending on the similarity metric, each have generated different synthetic samples which have been separately used for training the classifier.
An additional control parameter is the application of the similarity score either directly onto the time signals or onto their power spectral densities.
A summary of all variants are shown in Table~\ref{tab:classifier_model_summary}.
In total, classification results from $9$ different training sets have been evaluated.
For further simplifications, we use the preceding abbreviation "OT" (Optimal Control Training) for the training set which contains synthetic sequences that have been generated with the IMUDiffusion model according to Section~\ref{section:monitoring_ddpm_training}. If additionally the denoising process has been monitored, the abbreviation "OT-D" (Optimal Control Training with Denoising) is applied.

\begin{table}[h]
    \caption{
    Summary of the training sets that were used to the train the classifier.}
    \centering
    \scalebox{0.7}{
    \begin{tabular}{lcccccc}
        \hline
        \makecell{Training Set\\Name} & \makecell{Number of Real\\Training Samples} & \makecell{Synthetic \\Sequences Added} & \makecell{Number of\\Training Epochs} & \makecell{Number of\\Denoising Steps} & Similarity Metric & \makecell{Similarity Measured\\between} \\
        \hline \\
        2 Sample & 22 & None & 4500 & 3000 & None & None\\
        Full-Set & $\approx$ 25000 & None & 4500 & 3000 & None & None \\
        Full DDPM & 22 & 15360 & 4500 & 3000 & None & None \\
        OT C-Opt GAK & 22 & 15360 & Optimal Control & 3000 & C-Opt GAK & PSD\\
        OT-D C-Opt GAK & 22 & 15360 & Optimal Control & Optimal Control & C-Opt GAK & PSD\\
        OT Cosine PSD & 22 & 15360 & Optimal Control & 3000 & Cosine & PSD\\
        OT-D Cosine PSD & 22 & 15360 & Optimal Control & Optimal Control & Cosine & PSD\\
        OT Cosine Time & 22 & 15360 & Optimal Control & 3000 & Cosine & Time\\
        OT-D Cosine Time & 22 & 15360 & Optimal Control & Optimal Control & Cosine & Time \\
        \hline \\
    \end{tabular}
    }
    \label{tab:classifier_model_summary}
\end{table}

\section{Results}
The results section is divided in three parts.
The first two parts analyse the findings from integrating the similarity scores in the training and synthetization process of a diffusion model.
In the last part, the results of using  synthetic sequences for training a classification model are discussed. 

\subsection{Monitoring the DDPM Training Process}
\begin{figure}
    \centering
    \includegraphics[width=0.99\linewidth]{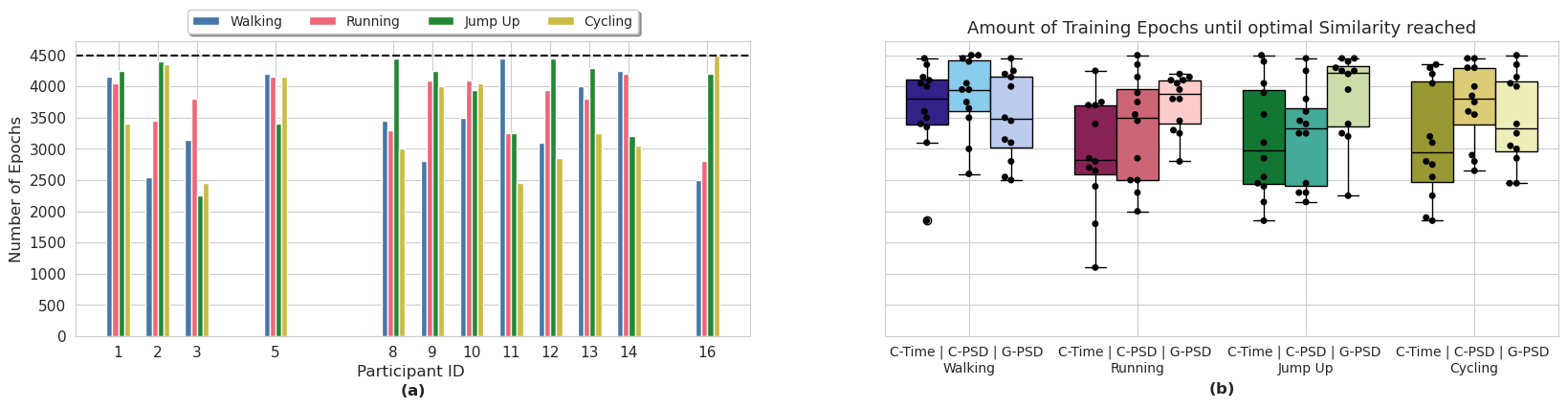}
    \caption{Amount of training epochs until the similarity score induced an early stopping of the training process. 
    Subgraph (a) visualizes the participant and class individual result when using the C-Opt GAK similarity metric for early stopping. 
    Subgraph (b) highlights the findings across all evaluated similarity metrics in a swarm-box-plot separated by the classes. 
    Each black dot in the swarm-plot represents one participant.
    }
    \label{fig:saved_ddpm_epochs}
\end{figure}

We have integrated the similarity scores in the training process of our IMUDiffusion model as some kind of early stopping criteria (OT-variants).
This allowed us to reduce the amount of training epochs.
Figure~\ref{fig:saved_ddpm_epochs} (a) visualizes the amount of training epochs until the training process has been terminated by this early stopping criteria.
It is shown separately for each participant and each activity that the participants performed.
In this graph, we only visualized the results that we obtained by using the C-Opt GAK similarity score calculated between the PSDs of the signals.
A summary across all three methods is visualized in Figure~\ref{fig:saved_ddpm_epochs} (b).
The results are further divided by the four activities.
In general, we can see a reduction in the amount of training epochs independent of the similarity score function used. 
With $1100$ training epochs, the fastest termination of the training occurred whilst monitoring the training process using the Cosine similarity metric applied to the signals in the time domain.
On average, this method required the least amount of training epochs until it stopped the training.
With it, we were able to reduce the amount of training epochs by $28.70\%$.
By using the Cosine similarity between the PSDs of the signals as metric reduced the amount of training epochs by $21.62\%$.
Finally, the C-Opt GAK similarity metric allowed us to reduce the amount training epochs the least with a reduction of $19.51\%$. 
A summary of the reduced amount of training epochs per class and similarity metric is given in Table~\ref{tab:saved_trainingepochs}.

\begin{table}
    \centering
    \caption{
    The average training epoch at which the similarity score reached a local optima and led to an early stopping of the training of the IMUDiffusion model. 
    Without this early stopping criteria, the amount of training epochs was set to $4500$.}
    \begin{tabular}{lcccc}
        \hline    
        Similarity Score & Walking & Running & Jump Up & Cycling\\
        \hline
        Cosine (PSD) & $3857.33 \pm 579.09$ & $3315.67 \pm 820.65$ & $3219.83 \pm 740.06$ & $3715.67 \pm 615.20$ \\
        Cosine (Time) & $3657.33 \pm 682.16$ & $2924.00 \pm 860.35$ & $3144.83 \pm 871.41$ & $3107.33 \pm 883.37$ \\
        C-Opt GAK (PSD) & $3507.33 \pm 664.84$ & $3744.83 \pm 426.96$ & $3861.50 \pm 659.58$ & $3457.33 \pm 696.97$ \\
        \hline
    \end{tabular}
    \label{tab:saved_trainingepochs}
\end{table}

\subsection{Monitoring the Denoising Process}\label{section:eval_denoising_process}
\begin{figure}
    \centering
    \includegraphics[width=0.9\linewidth]{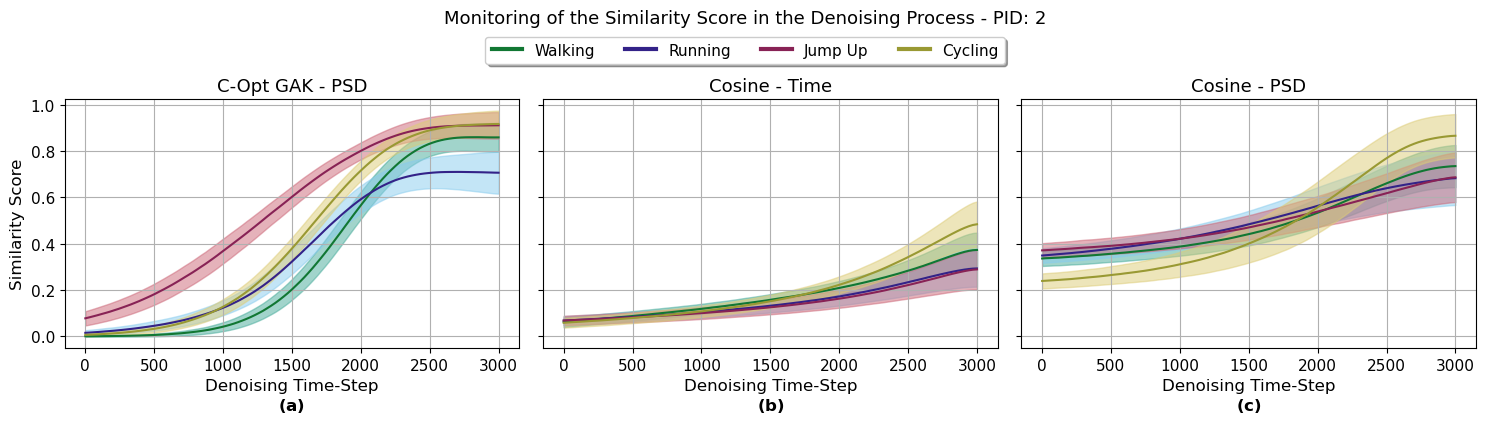}
    \caption{This graph visualizes the similarity scores for specific denoising steps in the denoising process for a single participant (PID 2).
    Subgraph (a) visualizes the C-Opt GAK score value whereas subgraph (b) and (c) visualizes the Cosine similarity score once between the signals in the time domain once between their PSDs.
    It is further separated by the four activities Walking, Running, Jump Up and Cycling.
    }
    \label{fig:all_denoising_processes}
\end{figure}

In this section, we describe the results from using the similarity score functions for monitoring the denoising process. 
An exemplary result of a monitored denoising process over one selected participant for all three metrics - C-Opt GAK, Cosine PSD and Cosine Time - is visualized in Figure~\ref{fig:all_denoising_processes}.
At the first denoising step (Denoising Time Step = 0), we start with a standard normal distributed signal.
Interestingly, the Cosine similarity between the PSDs of the gaussian white noise and real sequences showed some kind of similarity, see Figure~\ref{fig:all_denoising_processes} (c).
Even score values of around $0.4$ were reached. 
The score value did still increase with the denoising steps, though, in some cases only marginally from $0.4$ to around $0.6$.
This small increase of the score value could also be observed when the Cosine similarity was calculated between signals in the time domain, see Figure~\ref{fig:all_denoising_processes} (b). 
Though, this time, the score value started on average at around $0.064$ and did end with a similarity score value of approximately $0.484$ on average at the last denoising step.
This was at least the case for the Cycling class.
For the Jump Up class, the similarity score did on average reach a value of around $0.290$. 
Finally, the C-Opt GAK metric was able to broaden the range, see Figure~\ref{fig:all_denoising_processes} (a).
For example, on average a score of around $0.0070$ was reached at the first denoising step with the Cycling class. 
It did increase on average to $0.917$. \newline
Using the C-Opt GAK metric, highest score values were reached around the $2798^{th}$ denoising step on average.
When using the Cosine PSD and Cosine Time metrics, highest scores were reached around the $2885^{th}$ and $2933^{th}$ denoising step respectively.

\subsection{Classification Results}
\begin{figure}
    \centering
    \includegraphics[width=0.65\linewidth]{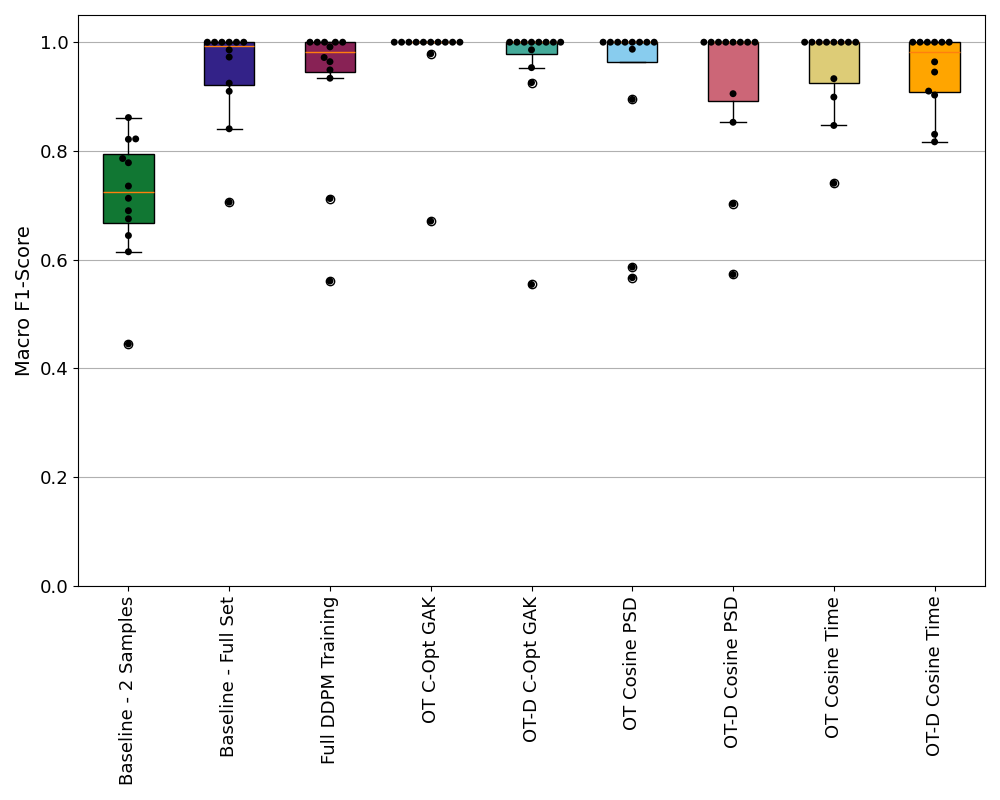}
    \caption{
    Participant individual macro F1-scores calculated with the nine classification models represented as swarm-boxplot where each dot represents one participant.
    Compared are the three results from the baseline models 2 Samples, Full-Set and Full DDPM Training against the results from the six models that were trained with synthetic sequences that were generated using the similarity metrics.
    The abbreviation OT stands for Optimal Control Training and the abbreviation D stands for Denoising. 
    }
    \label{fig:classification_results}
\end{figure}

The results with the baseline sets have already been discussed in detail in \cite{imudiffusion}. 
For comparison reasons, they are still added to the evaluation and graph that visualizes the classification results, see Figure~\ref{fig:classification_results}.
To be more specific, the graph visualizes the macro F1-scores across all test subjects according to the LOSOCV approach, and that individually for all $9$ training sets used to train the classifier.
The macro F1-scores are visualized as a swarm-box-plot, where each dot represents the score of one left out participant and each box depicts a statistical analysis over all those subject-individual results.
Best results were achieved with the OT C-Opt GAK set as with only two participants (PID 3 and 12) a macro F1-score of less than $1.0$ were reached.
With the remaining sets we achieved higher test scores for those two participants.
Except with the 2 Sample set, where the test score dropped even further.
Sequences from the Running and Walking activities were mixed up with each which led to the deterioration of the score value.
For PID 3, sequences from the Cycling class were also mixed up with sequences from the Jump Up class. 
\newline

\subsubsection*{Impact of using the similarity scores for early stopping in the denoising process of the DDPM}
This section analyses the results of using sequences for the classification task that were generated using an early stopping criteria within the denoising process according to Section~\ref{section:eval_denoising_process}.
Compared with the results obtained by using the two baseline sets 2 Samples and Full-Set for training the classifier, the early stopping of the denoising process led to sequences that mostly improved the classification task.
By using the C-Opt GAK set, the macro F1-score increased for all 12 participants when compared to the results of the 2 Sample set.
Against the results obtained by using the Full-Set, the macro F1-score improved for 4 participants and decreased for the participants 3 and 12.
Using the Cosine metric either in the time domain or by using the PSDs of the signals, the macro F1-scores improved for 4 participants.
In the same way did the score value decrease with 4, respectively 3 participants.\newline
Visually comparing the results obtained with the OT-D sets against the OT sets when the same metric was used for early stopping showed a decreasing performance for all sets evaluated, see Figure~\ref{fig:classification_results}.
Using the C-Opt GAK metric for monitoring indeed decreases the results for the participants 1 and 16.
In percentages, the scores decrease by $44.55\%$ and $7.43\%$ respectively.
Though, it also increases for the participants 3 and 12 by $28.22\%$ and $0.73\%$ respectively.
Analysing the results that have been  obtained using the Cosine metric, the additional early stopping of the denoising process improved the results for a single participant when the PSDs of the signals has been compared. It decreased for three participants.
In the time domain, the OT-D Cosine Time set improved the results of the classifier for three participants and decreased it for four participants.

\subsubsection*{Impact of calculating the similarity score in the time domain or between power spectral densities}
We have used the information of the Cosine similarity score for stopping the training of a diffusion model and its denoising process earlier than scheduled.
The similarity itself was calculated between signals in the time domain and their power spectral densities.
In this section, the classification results between those two approaches are presented.
When using the similarity score only for monitoring the training process, the OT Cosine PSD set led to an improvement of the  macro F1-score for two participants compared to the results obtained with the OT Cosine Time set. 
The scores improved once by $10.1\%$ and once by $15.3\%$.
In contrast to that, the classification results improved for four participants when trained with the OT-Cosine Time set.
In the best case, the macro F1-score increased by $36.6\%$ for PID 1 and the least improvement was achieved for PID 4 with an improvement of around $1.3\%$.
When using the similarity score also in the denoising process, we achieved a higher macro F1-score for four participants with an increase of up to $42.7\%$, whereas for three participants we achieved a lower score value with a reduction of less than $10\%$.

\section{Conclusion}
To rely on an objective criteria to supervise the training of a neural network classification model is a normal approach to stop the optimization process at the most beneficial timestep.
In contrast, the supervision of time series DDPMs is not as straightforward and mostly based on the knowledge of the user.
An objective criterion is missing in this field.
We tried to fill this gap by integrating existing and novel similarity score functions into the training and denoising process of a time series DDPM.
The novel similarity score function is based on an existing alignment function which we adapted to best fit the underlying dataset.
Therefore, we fit the similarity function to the training and validation sets by adjusting the scaling factor $\sigma$ of the initial similarity score function GAK.
The idea is to find an optimal $\sigma$ to assert high similarities between signals under the assumption that dissimilarities still exist within the sequences.
The adapted similarity score function is called C-Opt GAK.
The similarity metrics can then be used to not only monitor the diffusion models performance, but also for stopping the training and denoising as soon as an optimal similarity between real and synthetic sequences was achieved.
Generated sequences then have been used to train a classifier with the task of differentiating the four classes.
It served as an objective criteria for evaluating the effectiveness of using similarity scores to optimize the training and synthetization process of diffusion models. \newline
\newline
By using the C-Opt GAK metric for early stopping the training of the diffusion model, we were able to reduce the amount training epochs on average by $20\%$.
Across all participants and classes, this saved us $41,148$ epochs.
This not only saved computation time, but the classification results improved with six participants compared to when the classifier was trained with synthetic sequences that were generated after training the diffusion model for the full $4500$ epochs.
For one participant, the macro F1-score increased by up to $43.9\%$.
Independent of the similarity metric used for early stopping, the macro F1-score increased with more participants as it decreased, showing the effectiveness of this approach. \newline
\newline
Another approach is to integrate the similarity metrics in the denoising process.
This allows to stop the process as soon as highest similarity between sequences was reached.
Depending on the similarity metric, the score value itself could be misleading. 
The score values from the Cosine similarity between power spectral densities showed similarity values between $0.2$ and $0.4$ when real sequences have been compared against signals depicting random gaussian noise.
Nevertheless, the classification results improved with five participants compared to the results obtained with the Full DDPM set.
In contrast to that, the results dropped for either two or three participants depending on the similarity metric used.

\section{Discussion \& Outlook}
In this paper, we investigated the possibility of using similarity score functions to monitor the training and denoising process of a DDPM. 
The effectivity of those score functions was shown on a real world human activity recognition dataset.
We were able to reduce the amount of training epochs as well as denoising steps without missing out on the key characteristics that define the human activity it represents.
This could be verified by using those generated sequences to train a classifier.
For most LOSOCV steps, the additional synthetic sequences which were generated with the monitored DDPM improved the separability of the classifier.
Even though this was not the case with all participants.
With the help of the similarity score we were able to estimate the quality of the synthetic sequences, which resulted in the identification of sequences showing high dissimilarities.
It would be wise to integrate a selection process to identify the most suitable sequences improving the classifiers performance even further and reduce the required amount of synthetic sequences to a minimum. \newline
\newline
The monitoring of the denoising process including an early stopping criterion is a non-intuitive approach.
The diffusion model was trained in combination with a pre-chosen scheduler, which is in the generation process responsible for removing noise successively.
So, stopping the denoising process earlier is leading to sequences containing more shares of high frequency noise compared to sequences generated after the last denoising step.
We were still able to maintain the quality of the classifiers separability compared to the classifiers trained with the baseline training sets.
Though, it might affect different time series signals from other sensor types or other activities from the same sensor type differently.
So, it would be recommended to test this approach for different types of sensors and activities.
The advantage of requiring less denoising steps in the synthetization process is unambiguous.
It reduces the time to generate the sequences.
Additionally, it would be interesting to test different methods that reduce the amount of denoising steps for a DDPM against this approach. \newline
The range of the standard deviation for the calculation of the sigma value for the C-Opt GAK similarity score was chosen based on subjective criteria.
It would be interesting to analyse different ranges and its impact on different datasets. 
Especially for datasets with acyclic movements.

\bibliography{sample}

\section*{Acknowledgements}
This work has been partially funded by the Carl-Zeiss-Stiftung.

\section*{Author contributions statement}
Conceptualization, M.M.; Data curation, H.O.; Formal analysis, H.O. and M.M.; Funding acquisition, M.M.; Investigation, H.O., A.S. and M.M.; Methodology, H.O., A.S. and M.M.; Project administration, M.M.; Resources, M.M.; Software, H.O. and M.M.; Supervision, M.M.; Validation, H.O. and M.M.; Visualization, H.O.; Writing—original draft, H.O.; Writing—review and editing, A.S. and M.M. All authors have read and agreed to the published version of the manuscript.

\section*{Competing interests}
The authors declare no competing interests.

\section*{Data availability}
The data that support the findings of this study are available at the UCI Machine Learning Repository (https://archive.ics.uci.edu/dataset/305/realdisp+activity+recognition+dataset) and was originally collected by Oresti Banos et al.:
Banos, Oresti, Mate Toth, and Oliver Amft. "REALDISP Activity Recognition Dataset." UCI Machine Learning Repository, 2012, https://doi.org/10.24432/C5GP6D.

\end{document}